\documentclass[11pt, oneside, a4paper]{article}

\usepackage[cjk]{kotex}
\usepackage{setspace}
\usepackage[a4paper,left=20mm,right=20mm,top=30mm,bottom=30mm]{geometry}
\usepackage{tikz}
\usepackage[letterspace=200]{microtype}
\usepackage{graphicx}
\usepackage{amssymb}
\usepackage{endnotes}
\usepackage[normalem]{ulem}
\useunder{\uline}{\ul}{}

\usepackage{fancyhdr}
 
\begin{document}

\setstretch{1.2}

\title{Feature Extraction of Text for Deep Learning Algorithms: Application on Fake News Detection}

\author{HyeonJun Kim\medskip\\{\normalsize Soongsil University}}


\maketitle

\begin{abstract}
   
Feature extraction is an important process of machine learning and deep learning, as the process make algorithms function more efficiently, and also accurate. In natural language processing used in deception detection such as fake news detection, several ways of feature extraction in statistical aspect had been introduced (e.g. N-gram). In this research, it will be shown that by using  deep learning algorithms and alphabet frequencies of the original text of a news without any information about the sequence of the alphabet can actually be used to classify fake news and trustworthy ones in high accuracy (85\%). As this pre-processing method makes the data notably compact but also include the feature that is needed for the classifier, it seems that alphabet frequencies contains some useful features for understanding complex context or meaning of the original text.\\\\

\textbf{keywords:} {[FEATURE EXTRACTION], [DEEP LEARNING]}


\end{abstract}

\newpage

\tableofcontents

\newpage
\setstretch{1.8}



\section{Introduction}

Natural language processing for deception detection(e.g. fake news detection) focus on preprocessing text into computational data with required features for the propose. As deception detection is about understanding the meaning of the text or how the text is viewed by people, the sequence of the text is always considered as one of primary source of context. For example, N-gram, the representative method of natural language processing, contains the data of a word and its subsequent word and its statistical probabilities. The attribute \textit{subsequent} contains the continuous context of the text, or the linguist  will describes as linearity. In contrast, feature extractions without considering this language's linearity seems to be nonsense. However, if the data that been processed out of those non-linear feature extractions shows notable accuracy of detecting deceptions, it is possible to suggest that some of those preprocessing methods could be used as one of possible natural language processing for certain situations. \

In this paper, we discuss the effectiveness of APV, a simple natural language processing method using alphabet frequency, in the context of application on fake news detection. By using deep learning algorithm and fake news dataset in Kaggle, our findings suggest that simple deep learning algorithms using APV as pre-processing method could show prominent accuracy on predicting deception of the text. \

In section 2, we investigate conventional natural language processing that is used for machine learning and deep learning algorithms. In section 3, we define APV and its mathematical structure. We will also discuss the hypothesis that might improve feature extraction of APV.  In section 4, basic experiment protocol will be set including the structure of deep learning algorithms and performance metrics that will be used in the experiment. In section 5, we present the result of the algorithms performance. Finally, in section 6, we conclude the study.

\section{Previous Works}
\label{prework}

Most of previous works generally focus on preprocessing that uses statistical frequency of the words of text or social context of the text. For example, there is a network approach using network data of social media \cite{yang2019unsupervised}, or Term Frequency-Inverse Document Frequency(TF-IDF) of bigram \cite{gilda2017evaluating} as frequency approach. And considering all of these results, there is a approach to combine the data of the text of a news article, people's response to the article, and the article's source \cite{ruchansky2017csi}.\

In the aspect of prediction algorithms using those pre-processed data, there has been attempts to use deep learning algorithms. For example, here is a approach using network data and geometric deep learning, which achieved 92.7\% accuracy \cite{monti2019fake}. There is also a comprehensive report of accuracy using different N-gram models, TF-IDF or TF, and conventional machine learning algorithms, which some of them shows notable accuracy(92\%) using 50,000 features of TF-IDF, uni-gram \cite{ahmed2017detection}. 

\subsection{Discussion of Previous Works and Setting Goal for the study }

It is not clear how much features are needed to detect deception of the text. However commercially speaking, it is more convenient if the pre-processing effort is smaller, and still can retain a similar accuracy. Therefore, to be a competent model, a natural language processing model should be easier to apply and cost less time and space complexity, and at the same time have similar accuracy as other previous works, without depending on sophisticated algorithms for predictions.

\section{Proposed Methods}
\label{apv}

In this section we introduce simple natural language processing method \textit{Alphabet Probability Vector (APV)} and the way to improve its feature extraction.

\subsection{Definition of APV}

Alphabet probability vector is a 26 dimensional vector that contains the value of each alphabet frequency divided into sum of the frequency. $$v_{apv} = \left[ \ p_1 \ p_2 \ \cdots \ p_{26} \ \right]$$

$$p_k = {n_k \over \sum_{k} n_k}, \ n_k=\mathrm{(frequency \  of\  }k \mathrm{th \ alphabet)}$$The order of the alphabet data is same as it is commonly used.
As the APV varies depending how long is the text (denominator) and the frequency of the alphabets (numerator), The full text should be used for evaluation, i.e. not intentionally set to same denominator between APVs, because that will likely cause the loss of letters which is important to evaluate the content.\ 

The limitation of APV is that it cannot distinguish anagrams, however this usually doesn't matter because APV will be used for large-sized text evaluations, thus it is highly unlikely for two completely different text to have same APV. However deleting these problem-making data could improve the testing ability. 

\subsection{Improving of APV Feature Extracting}

By experimentation, it becomes clear that setting a positive integer $N$ to frequency of unused alphabets (N-Supplied APV, N-SAPV) actually improve the accuracy. We will discuss about this phenomenon soon.\

Also, it could be suggested that the APV can be more explicitly distinguished by its difference between APV of commonly used words. Department of mathematics, Cornell University had shared the normalized alphabets frequency of 40,000 English words\footnote{http://pi.math.cornell.edu/~mec/2003-2004/cryptography/subs/frequencies.html}. We defined a APV based on this data ($v_0$) , and added another process that substracts text APV with $v_0$. $$v_{apv}' = v_{apv} - v_0$$ This model will be called Standard-Subtraction Modification (SSM) in this paper.

\section{Experiment Protocol}
\label{protocol}
The experiment is designed using Python 3.8 and Keras 2.4.0, running on macOS environment. Firstly, the dataset is obtained, then by using APV or N-SAPV, and using or not using SSM, the dataset is pre-processed and been fed to Deep Neural Network(DNN). To observe feature extraction performance of N-SAPV as the N goes up, another special experiment will be designed to obtain the observation. Although the goal was detecting deception, we are using supervised learning method to check APV's feature extraction ability.

\paragraph{Preparing Dataset}

The dataset that been used in this experiment is the copy of the dataset used in other previous studies \cite{ahmed2017detection} \cite{ahmed2018detecting}. This dataset is easily downloadable in the Internet site Kaggle. The dataset consist of 44,898 rows, news title and text as columns. The fake news data are separated in different file, so the data was labeled (fake: 0, true: 1) and then merged to one dataset. The dataset is divided by training set and testing set, which are each random samples of 30,000 rows and 9,898 rows without any common row.

\paragraph{Preprocessing}

The dataset was pre-processed according to APV, 1-SAPV. For comparison, each situation has modified version. The modification had followed SSM. By defining a python object that counts the frequency of the alphabets and normalized it with sum of frequencies. There are other type of preprocessing concerning possible $N$ of N-SAPV.

\paragraph{Deep Neural Network Design}

Using Keras, we designed a simple DNN with one input layer, two hidden layers and one output layer. The input layer and hidden layers have 128 nodes with ReLU as activation function, and a output layer have one node with Sigmoid as activation function. For loss function, Binary Cross Entropy Loss is used without smoothing.

\paragraph{Performance Metrics}

For performance metric, we use accuracy, precision, recall, and F$_1$-score. These metrics are broadly used for performance evaluation of machine learning classifier \cite{yang2019unsupervised}.

\section{Experiment Results}
\label{result}

First set of results is based on DNN trained for 600 epochs and 200 batch size, which is shown in Table \ref{table:1}. The results shows that unmodified APV has more unbalanced error rates. N-SAPV seems to solve this problem, and SSM seems to solve the problem as well.

\begin{table}[h!]
\begin{center}

 \begin{tabular}{||c c c c c||} 
 \hline
 Type & Accuracy & Precision & recall & F$_1$-score\\ [0.5ex] 
 \hline\hline
 APV &0.78 &0: 0.72, 1: 0.91&0: 0.94, 1: 0.60 &0: 0.82, 1: 0.73 \\ 
 \hline
 APV, SSM &0.83 &0: 0.85, 1: 0.81&0: 0.82, 1: 0.84 &0: 0.83, 1: 0.83\\
 \hline
 1-SAPV &0.84 &0: 0.84, 1: 0.84&0: 0.85, 1: 0.82 &0: 0.84, 1: 0.83\\
 \hline
 1-SAPV, SSM &0.84 &0: 0.85, 1: 0.84&0: 0.86, 1: 0.83 &0: 0.85, 1: 0.83\\
 \hline

\end{tabular}

\caption{Performance metrics of DNN using each method of pre-processing.}
\label{table:1}
\end{center}
\end{table}

Second set of results in Table \ref{table:2} shows the accuracy of N-SAPV, SSM, DNN of 1000 epochs and 200 batch size depending on $N$. According to the results, There is no significant growth of accuracy for N over 2. This might implies N over 2 will be not effective, which means choosing N larger than 0 should be considered with care as it modifies the original data.

\begin{table}[h!]
\begin{center}

 \begin{tabular}{||c || c c c c c c||} 
 \hline
  &0-SAPV (APV)&1-SAPV &2-SAPV &3-SAPV &4-SAPV&5-SAPV\\ [0.5ex] 
 \hline\hline
 Accuracy &0.83&0.84&0.85&0.84&0.85&0.84 \\ 
 \hline

\end{tabular}

\caption{Accuracy of DNN using N-SAPV, SSM depending on size of N}
\label{table:2}
\end{center}
\end{table}
\section{Conclusion}

As the current natural language processing suggests, it is ideal for the natural language to be considered by their strict and linear order. However, in this paper, we suggest that even if the natural sequence of the data are excluded from the feature extraction of a text, it is possible to use those data to classify the text with high accuracy(85\%). And we consider this accuracy is a fair trade-off between accuracy and pre-processing effort, as APV shrinks the data to approximately ${26 \over n}$ ($n$ as about the whole length of text) in its size but still obtained 92\% of the accuracy the deep learning algorithm that is been reported has \cite{monti2019fake}.  It is not sure whether or not APV is capable of summarizing the text, however it seems possible to use APV in supervised learning regarding natural language. We are planning to use the proposed method to classify more than 2 classes, and also hopefully find mathematical explanations why this method works and how to improve the feature extraction more than the results listed in this paper.

\section{Acknowledgement}
The author wants to thank Prof. Kwang Baek Kim, Silla University, Republic of Korea for supporting and advising in the process of paper writing.

\vskip 0.2in

\bibliography{main}
\bibliographystyle{plain}
\end{document}